\documentclass[letterpaper, 10 pt, conference]{ieeeconf}  

\IEEEoverridecommandlockouts                              

\usepackage{epsfig} 
\usepackage{amsfonts}
\usepackage{algorithm} 
\usepackage{algpseudocode} 
\usepackage{comment}
\newtheorem{lem}{Lemma} 
\usepackage{textcomp}
\usepackage{multirow}
\usepackage{amsmath, xparse}
\usepackage{graphicx}

\usepackage[backend=bibtex,style=numeric,natbib=true,style=ieee]{biblatex}
\title{\LARGE \bf
Value Summation: A Novel Scoring Function for MPC-based Reinforcement Learning
}
\author{Mehran Raisi$^{1}$, Amirhossein Noohian$^{2}$, Luc Mccutcheon$^{1}$, and Saber Fallah$^{1}$
\thanks{}
\thanks{$^{1}$Mehran Raisi, Luc Mccutcheon, and Saber Fallah are with the Connected Autonomous Vehicle Lab (CAV-Lab), University of Surrey,
        Guildford, England
        {\tt\small [m.raisi, lm01065@surrey, s.fallah]@surrey.ac.uk}}%
\thanks{$^{2}$Amirhossein Noohian is with the Department of Mechanical Engineering, Sharif University of Technology,
        Tehran, Iran
        {\tt\small amirhossein.noohian@mech.sharif.edu}}%
}

\begin{document}

\maketitle
\thispagestyle{empty}
\pagestyle{empty}

\begin{abstract}
This paper proposes a novel scoring function for the planning module of MPC-based reinforcement learning methods to address the inherent bias of using the reward function to score trajectories. The proposed method enhances the learning efficiency of existing MPC-based MBRL methods using the discounted sum of values. The method utilizes optimal trajectories to guide policy learning and updates its state-action value function based on real-world and augmented onboard data. The learning efficiency of the proposed method is evaluated in selected MuJoCo Gym environments as well as in learning locomotion skills for a simulated model of the Cassie robot. The results demonstrate that the proposed method outperforms the current state-of-the-art algorithms in terms of learning efficiency and average reward return.
\end{abstract}


\section{Introduction}
Reinforcement learning (RL) has recently received great attention from robotics and autonomous systems communities due to its power in handling system non-linearity, uncertainties, and complexities. The current RL methods can be classified as Model-free RL (MFRL) and Model-based RL (MBRL). Although MFRL methods have demonstrated proficiency in complex tasks \cite{c30,c31}, they are typically sample-inefficient and suffer from a prolonged training process. In contrast, MBRL methods are usually more sample-efficient, since they learn the dynamic transition model and can use it for planning. In order to learn the model of the environment, methods such as ensemble learning \cite{c3} and latent models \cite{c4} have been proposed, allowing for fast and robust learning of the environment dynamics. Once the model has been obtained, it can be used together with a planner to achieve a wide variety of tasks and goals.\\
MBRL planner algorithms can be categorized as background planning and discrete-time planning \cite{c7}. The former uses the transition data obtained from the model to improve the policy or to learn the value function, while the latter mainly performs planning online for every state that the agent visits. Background planning algorithms include dynamic programming \cite{c8}, tabular Dyna \cite{c9}, and prioritized sweeping \cite{c10}. Examples of discrete-time planning are Monte Carlo Tree Search (MCTS) \cite{c19} and Model Predictive Control (MPC) \cite{c20}. MCTS algorithms showed superior performance in discrete spaces such as Atari games, while MPC algorithms are mainly used for continuous applications such as robotics. POLO is a prominent example of MPC-based MBRL, proposed by Lowrey et al., in which the concepts of value function approximation are combined with MPC to stabilize and accelerate the learning process of the value function \cite{c21}. The work on POLO was extended by Morgan et al. \cite{c22} through modeling MPC as an actor-critic framework and incorporating the approximation error when learning the dynamics model. Another approach, given by Sikchi et al. \cite{c32} integrated the value function learned from model-free off-policy RL into MPC, and proposed actor regularized control to address the issue of actor divergence. Hoeller et al. \cite{c34} defined the objective function of MPC in terms of an estimated value function and showed that MPC minimizes an upper bound of the cross-entropy sampling method to the state distribution of the optimal sampling policy. Zong et al. \cite{c33} combined the concepts of iLQR control with MFRL, resulting in an algorithm with low computation overhead, high sample efficiency, and robustness against the errors of the model. Hansen et al. \cite{c35} used a learned task-oriented dynamics model for MPC and a learned terminal value function, where both the model and the value function are learned jointly by temporal difference learning. The concepts of Model Predictive Path Integral Control (MPPI) and MFRL were combined by Charlesworth et al. \cite{c36}, resulting in a high-performance algorithm in some challenging simulated manipulation tasks where current RL methods and MPC techniques perform poorly.\\
Typically, MPC methods are based on a random shooting process. This process samples numerous trajectories composed of state-action sequences with $H$ horizon and evaluates each trajectory with a scoring function \cite{c20}. The defined scoring functions in literature can be classified into two types. The first one sums the immediate rewards along the trajectory to calculate the trajectory score $\sum_{t=0}^{H} \gamma^{t}r(s_{t}, a_{t})$ \cite{c3, c11}, where $\gamma \in [0, 1)$ is the discount factor which balances the emphasis on imminent rewards over distant rewards. The second one sums immediate rewards along the trajectory and considers the estimated value of the terminal state as the terminal reward $\sum_{t=0}^{H-1} \gamma^{t}r(s_{t}, a_{t}) + \gamma^{H}\hat{V}(s_{H})$ \cite{c21, c22, c32, c33, c34, c35, c36}. However, using immediate rewards within the trajectory may not be enough for action selection, since no information about the future is considered. When operating in environments with sparse reward signals, utilizing immediate rewards within the planning horizon may lead to indistinguishable scores across all trajectories \cite{c6}. Additionally, in complex environments, where states are densely distributed and sampling time is constrained, immediate rewards may not be effective in distinguishing between different states.\\
In order to address the aforementioned problems, we propose the use of the discounted summation of estimated state-action values as the scoring function for MPC-based MBRL algorithms, where the score of a trajectory is calculated by summing state-action values rather than rewards. It should be noted that we use state-action values within the planning horizon instead of state values since state-action values pertain to the action taken in the trajectory, making the score accurate within the context of the trajectory. Moreover, summing rewards to score trajectories emphasizes actions with greater immediate reward, whereas summing state-action values emphasizes actions with greater cumulative rewards, thereby increasing total return over the episode. The results show that the proposed method improves learning efficiency and average return when compared to state-of-the-art baselines. To the best of our knowledge, it is the first time a discounted sum of values has been used in MBRL systems.\\
The remainder of the paper is structured as follows: we discuss the related works in Section II, while Section III describes preliminaries for key components of MBRL. Section IV introduces the value summation function and its integration in MBRL. The results are presented in Section V. Finally, our contributions and future works are summarized in Section VI.
\section{Preliminaries}

\subsection{Reinforcement Learning}
In the context of RL, a problem can be formulated as a finite-horizon Markov Decision Process (MDP) $\mathcal{M = <S, A, R, P, \gamma>}$, where $\mathcal{S}$ denotes the state space, $\mathcal{A}$ denotes the action space, $\mathcal{R: S \times A} \rightarrow \mathbb{R}$ denotes the reward function, $\mathcal{P: S \times A \times S} \rightarrow \mathbb{R}$ is the transition model, and $\gamma \in [0,1)$ is the discount factor. A policy $\pi \in \Pi: \mathcal{S \times A} \rightarrow \mathbb{R}$ describes a mapping from states to actions. In RL problems, return is defined as the accumulative discounted reward, given by ${G} = \sum_{t=0}^{\infty} {\gamma^t r(s_t,\pi(s_t))}$. An MDP aims to find the optimal policy $\pi^*$ that maximizes the expected return.\\
$V(s)$ denotes the state value, which is defined as the expected return by following the policy $\pi$ from the state $s$:
\begin{equation}
\label{eq1}
V(s) = \mathbb{E}_{\pi} \left[ \sum_{t=0}^{\infty} {\gamma^t r(s_t,\pi(s_t))} \right], s_0=s
\end{equation}
Similarly, $Q(s,a)$ denotes the state-action value, which is defined as the expected return by taking the action $a$ at the state $s$ and then following the policy $\pi$:
\begin{equation}
\label{eq2}
{Q(s,a)} = \mathbb{E}_{\pi} \left[ \sum_{t=0}^{\infty} {\gamma^t r(s_t,\pi(s_t))} \right], s_0=s, a_0=a
\end{equation}
In continuous MDPs, the policy and the state-action value function are parameterized as $\pi_{\theta}$ and $Q_{\phi}$, where $\theta$ and $\phi$ denote the vector of parameters. The parameters of the state-action value function $Q_{\phi}$ can be estimated using the following loss function:
\begin{equation}
\label{eq3}
J_{\phi} = \mathbb{E}_{s,a,r,s' \sim \mathcal{D}} \left[{(r + {\gamma} {\max_{a'} {Q_{\phi}(s',a')}} - Q_{\phi}(s,a))^2} \right]
\end{equation}
where $\mathcal{D}$ is a dataset containing a series of transitions $(s,a,r,s')$ collected by the policy $\pi_{\theta}$. After the state-action value function is approximated, the parameters of the policy are updated using the following function:
\begin{equation}
\label{eq4}
\nabla_\theta {J_{\theta}} = \mathbb{E}_{\pi_\theta} \left[ {\nabla_\theta \log{\pi_\theta(s,a)} Q_\phi(s, a)} \right]
\end{equation}

\subsection{MPC-based Reinforcement Learning}
MBRL approaches benefit from the model of the environment to plan the best action sequence at each state. In the context of MBRL, the transition model can be known or learned. When the transition model is available, it is used for planning the action sequence at each time step. In this regard, MPC is mostly used to plan the best action sequence. The MPC policy $\pi_{MPC}$ computes the best action sequence at each state using the transition model in three stages: trajectory sampling, trajectory evaluation, and action selection. In general, an MPC policy is computed as follows:
\begin{equation}
\begin{aligned}
\label{eq5}
\pi_{MPC}(s) = \arg \max_{a_{0:H-1}} {\mathbb{E} \left[ \sum_{t=0}^{H-1} {\gamma^t r(s_t,a_t) + \gamma^H r_f(s_H)} \right]},\\
a_t = \pi_t(s_t), s_0=s
\end{aligned}
\end{equation}
where states evolve according to the on-board model $s_{t+1}=\hat{f}(s_t,a_t)$, $\pi_t$ is a distribution for sampling action at each state, and $r_f$ is the terminal reward. After the optimal action sequence is computed, the first action is executed in the environment and the procedure is repeated at the next time step.
It is worth mentioning that, in some cases where MBRL is combined with MFRL, the distribution $\pi_t$ can be the parameterized policy $\pi_\theta$, called the behavioral policy.

\section{Proposed MBRL Method}
The proposed Value Summation MBRL (VS-MBRL) method is presented in this section. The concept of discounted value summation and its boundedness over the infinite horizon is explained. Then, the integration of the proposed scoring function in trajectory evaluation and the corresponding VS-MBRL is discussed.
\subsection{Scoring Function}
The use of reward-based scoring functions to evaluate trajectories and select the best action out of different trajectories can be misleading. The problem with the application of reward-based scoring functions is that the discounted summation of finite-horizon immediate rewards can be significantly biased against future decisions because the impact of future rewards is not considered in the scoring function. Even the selection of greater horizons $H$ cannot alleviate this problem, because they may impose cumulative errors and reduce planning efficiency. Another problem regarding the application of reward-based planning is that in environments with sparse reward feedback, where the total reward is received upon the completion of an episode, the summation of rewards within a planning horizon can be the same across all trajectories \cite{c6}. Last but not least, a problem occurs in environments with short sampling times, where the states' difference across each increment of time is too short to have a considerable impact on the reward function $r(s_{t}, a_{t})$. For these reasons, incorporating the estimated state-value function in the scoring function can be more insightful. To address these issues, we propose the discounted summation of estimated state-action values to score trajectories as:
\begin{equation}
\label{eq6}
S = \sum_{t=0}^{H}\gamma^{t}{Q^{\pi}}(s_{t},a_{t})
\end{equation}
where the estimated state-action values over a trajectory with the planning horizon $H$ are summed up to determine the trajectory score $S$. Using the Bellman equation, we can roughly estimate the state-action value of each time-step as follows:
\begin{equation}
\label{eq7}
Q^{\pi}(s_{t},a_{t}) = r(s_{t},a_{t}) + \gamma Q^{\pi}(s_{t+1},a_{t+1})
\end{equation}
Using \eqref{eq7} iteratively for all time steps within the planning horizon, we can write the scoring function in \eqref{eq6} in terms of the immediate rewards as below:
\begin{equation}
\begin{aligned}
\label{eq8}
S = & \sum_{t=0}^{H}(t+1)\gamma^{t}r(s_{t}, a_{t}) + \sum_{t=H+1}^{T}(H+1)\gamma^{t}r(s_{t}, a_{t})
\end{aligned}
\end{equation}
See Appendix A for a detailed derivation of (8). The obtained scoring function~\eqref{eq8} means the rewards within the planning horizon are emphasized by $(t+1)\gamma^{t}$ factor, and the rewards outside of the planning horizon are discounted by $(H+1)\gamma^{t}$. Compared to the conventional reward scoring function $\sum_{t=0}^{H} \gamma^{t}r(s_{t}, a_{t})$, future rewards have greater coefficients and more impact on the score of trajectory. This highlights the potential of the proposed scoring function in better emphasizing future rewards within the prediction horizon. We demonstrate that the proposed scoring function~\eqref{eq8} will not violate the boundedness of summation, even in an infinite-horizon ($H=\infty$) planning case (\textit{See} Lemma 1).
\begin{algorithm}
\caption{Value Summation MBRL}
\label{alg:VS-MBRL}
\begin{algorithmic}[1]
\State Given $f(s_t,a_t)$: real environment, $\hat{f}(s_t,a_t)$: on-board model, $D_{1}$: on-board replay buffer, $D_{2}$: environment replay buffer, $\pi_{\theta}(s_{t})$: actor-network, $Q_{\phi}(s_{t},a_{t})$: critic network, $\gamma$: discounted factor;
\State ${H}$: Horizon length;
\State ${N}$: Number of trajectories;
\While{Task Not Completed}

    \State $s_{t} \gets ReadSensors()$
    \For{$n \gets 0$ to $N$}
        \State ${s_{0}} = s_{t}$
        \State $S(n) = 0$
        \For{$h \gets 0$ to $H$}
            \State $a_{h} = \pi_{\theta}(s_{h})$
            \State $s_{h+1} = \hat{f}(s_{h}, a_{h})$
            \State $q_{h} = Q_{\phi}(s_{h}, a_{h})$
            \State $S(n) += \gamma^{h}q_{h}$
            \State $Store$ $(s_{h}, a_{h})$ $in$ ${D}_{1}$
        \EndFor
        \State $Use$ ${D}_{1}$ $to$ $Update(Q_{\phi})$
    \EndFor
    \State $a_{t} = \underset{a_{t:t+H}}{\mathrm{argmax}}\ S$
    \State $s_{t+1} = {f}(s_{t}, a_{t})$
    \State $Store$ $(s_{t}, a_{t})$ $in$ ${D}_{2}$
    \State $Use$ ${D}_{2}$ $to$ $Update(\pi_{\theta}, Q_{\phi})$

\EndWhile
\end{algorithmic}
\end{algorithm}

\begin{lem}
\label{Lemma 1}
Consider an infinite horizon action sequence ${A}= \{a_{0},a_{1},a_{2}, ...\} $ and its resultant reward sequence ${R} = \{r_{0}, r_{1}, r_{2}, ... \}$. With these assumptions, \eqref{eq8} is expressed as below:
\begin{equation}
\label{eq9}
S = \sum_{t=0}^{\infty}(t+1)\gamma^{t}r(s_{t},a_{t})
\end{equation}
Considering the maximum reward from the infinite reward sequence $R$ as $r_{max} = max \{r(s_{0},a_{0}), r(s_{1},a_{1}), r(s_{2},a_{2}), ... \}$, the scoring function is upper bounded by:
\begin{equation}
\begin{aligned}
\label{eq10}
\sum_{t=0}^{\infty}(t+1)\gamma^{t}r(s_{t},a_{t}) & \leq \sum_{t=0}^{\infty}(t+1)\gamma^{t}r_{max}\\
& = r_{max}\sum_{t=0}^{\infty}(t+1)\gamma^{t}
\end{aligned}
\end{equation}
where $\sum_{t=0}^{\infty}(t+1)\gamma^{t}$ has an analytical solution as follows (\textit{See} Appendix B):
\begin{equation}
\label{eq11}
\sum_{t=0}^{\infty}{(t+1)\gamma^{t}} = \frac{1}{(1-\gamma)^{2}}
\end{equation}
Substituting \eqref{eq11} into \eqref{eq10}, we have the following relation for the scoring function:
\begin{equation}
\label{eq12}
\sum_{t=0}^{\infty}(t+1)\gamma^{t}r(s_{t},a_{t}) \leq \frac{1}{(1-\gamma)^{2}}r_{max}
\end{equation}
Therefore the infinite horizon scoring function has an upper bound.
\end{lem}

\subsection{Value Summation MBRL}
\begin{figure}
\begin{center}
\includegraphics[width=8.5cm]{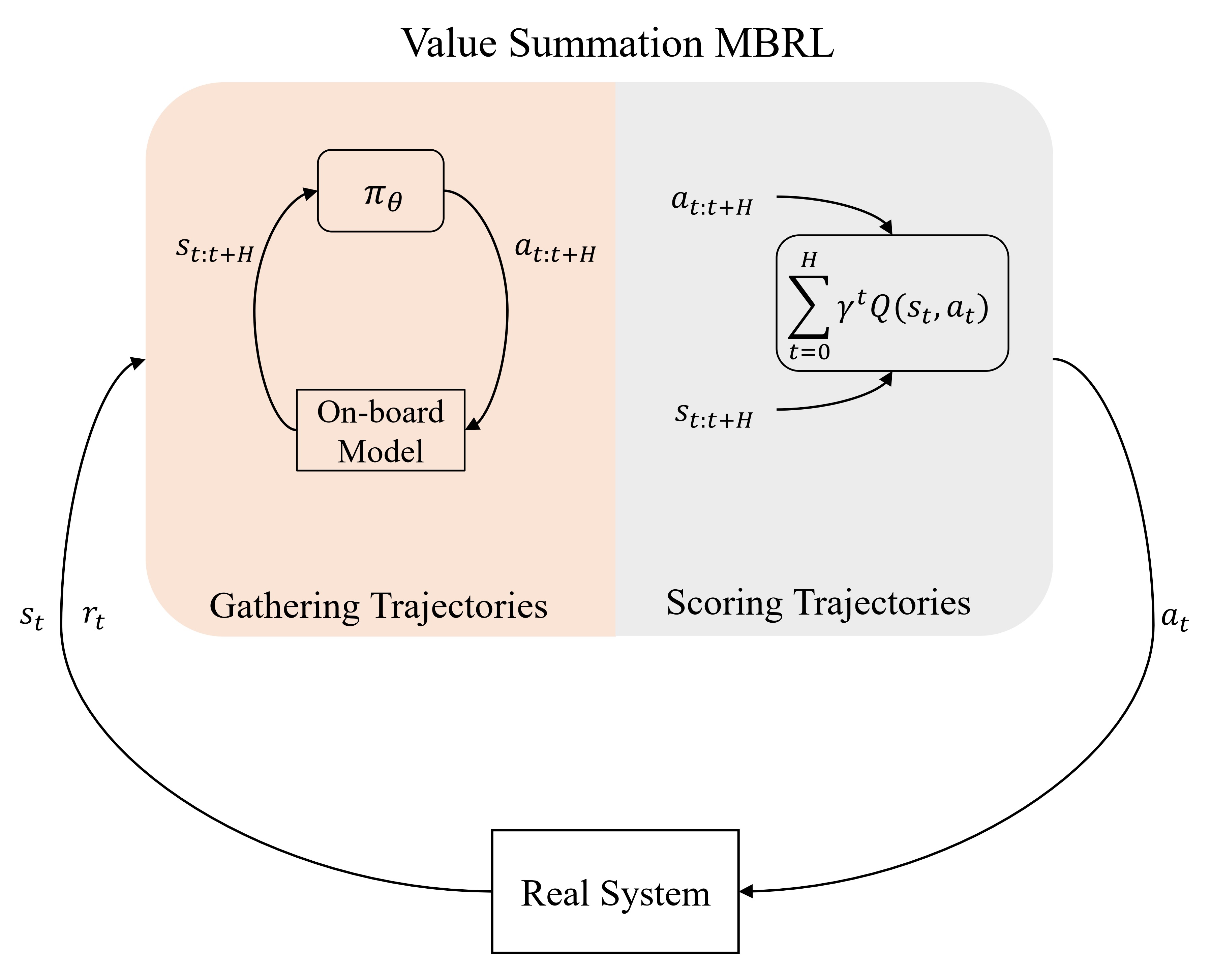}
\caption{The schematic of the Value Summation MBRL algorithm. The policy generates several trajectories, and the scoring mechanism selects the optimal trajectory with respect to the value summation function.}
\label{fig2}
\end{center}
\end{figure}

This subsection presents the VS-MBRL method, which integrates the value summation scoring function into planning (Fig.~\ref{fig2}). This algorithm consists of two sections: gathering trajectories and scoring them. After receiving the current state $s_{t}$ and the instant reward $r_{t}$ from the environment, the on-board model $\hat{f}(s_{t}, a_{t})$ is used to sample several trajectories with the planning horizon $H$ from the policy $\pi_{\theta}$. The collected trajectories, then, are scored by the value summation function, and the first action from the highest-scored trajectory is executed in the environment.\\
VS-MBRL employs Soft Actor-Critic (SAC) \cite{c14} as the underlying actor-critic algorithm to take advantage of the exploration induced by soft policy updates based on the maximum entropy principle. This property makes SAC an excellent candidate for MBRL applications since actions can be sampled directly from the policy $\pi_{\theta}$, and no additional noise is required. However, VS-MBRL can be applied to any developed off-policy actor-critic algorithm. A soft policy evaluation step is conducted in SAC using the soft Bellman backup operator \cite{c14}, followed by a soft policy improvement step in which the expected KL divergence is minimized.\\
A description of the VS-MBRL algorithm can be found in Algorithm ~\ref{alg:VS-MBRL}. Using the on-board model $\hat{f}(s_{t}, a_{t})$ with the planning horizon $H$, the agent shoots a number of trajectories at each time step (line code: $6-17$). Eq. (\ref{eq6}) is applied to evaluate each pair $(s_{t:t+H}, a_{t:t+H})$ (line code: $12-13$) such that the corresponding trajectory can be evaluated. The augmented data is collected in the replay buffer ${D}_{1}$, which is then applied to train the state-action value function (line code: $16$). The first action from the best-scored trajectory is then executed in the environment (line code: $19$). To reduce computation costs, the augmented data is only employed for updating state-action value functions, while the actor-network is updated by $1/{(N+1)}^{th}$ that of the critic network, where $N$ is the number of trajectories (line code: $21$).
\section{Simulation}
This section compares the proposed value-summation scoring function with the most recent proposed scoring functions in the literature. The scoring functions used for comparison purposes are:\\
1) \textbf{Sum-Reward}: Discounted sum of rewards $\sum_{t=0}^{H} \gamma^{t}r(s_{t},a_{t})$, applied in \cite{c3}.\\
2) \textbf{Sum-Reward-Value}: Discounted sum of rewards added with the estimated value of the terminal state, given by $\sum_{t=0}^{H-1} \gamma^{t}r(s_{t},a_{t}) + \gamma^{H}{Q}^{\pi}(s_{H}, a_{H})$, applied in \cite{c22}.\\
3) \textbf{Sum-Value}: Our proposed discounted sum of state-action values $\sum_{t=0}^{H} \gamma^{t}Q^{\pi}(s_{t},a_{t})$.\\
First, the performance of VS-MBRL is evaluated and compared in standard Gym environments. Then, VS-MBRL is integrated into the process of learning locomotion skills for the Cassie robot and its performance is compared with other baselines. In all simulations, neural networks have the same architecture, including two hidden layers with 256 neurons each. In addition, the ReLU activation function is selected for both layers and the Adam optimizer \cite{c40} is used to update the parameters of the network. The results indicate that the proposed scoring function outperforms other scoring functions in terms of learning efficiency and average return.
\begin{figure*}
\begin{center}
\includegraphics[width=17.5cm]{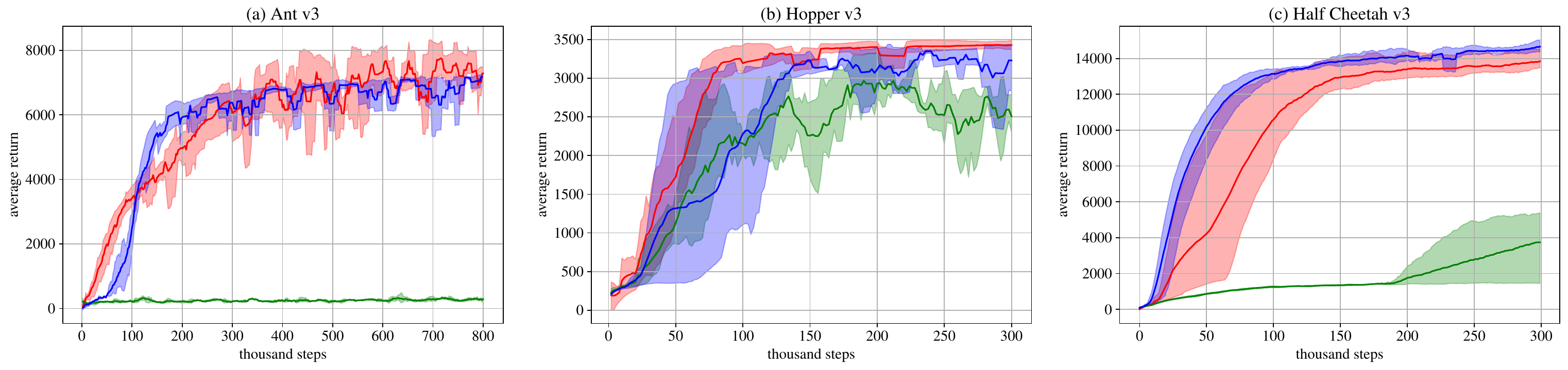}
\includegraphics[width=17.5cm]{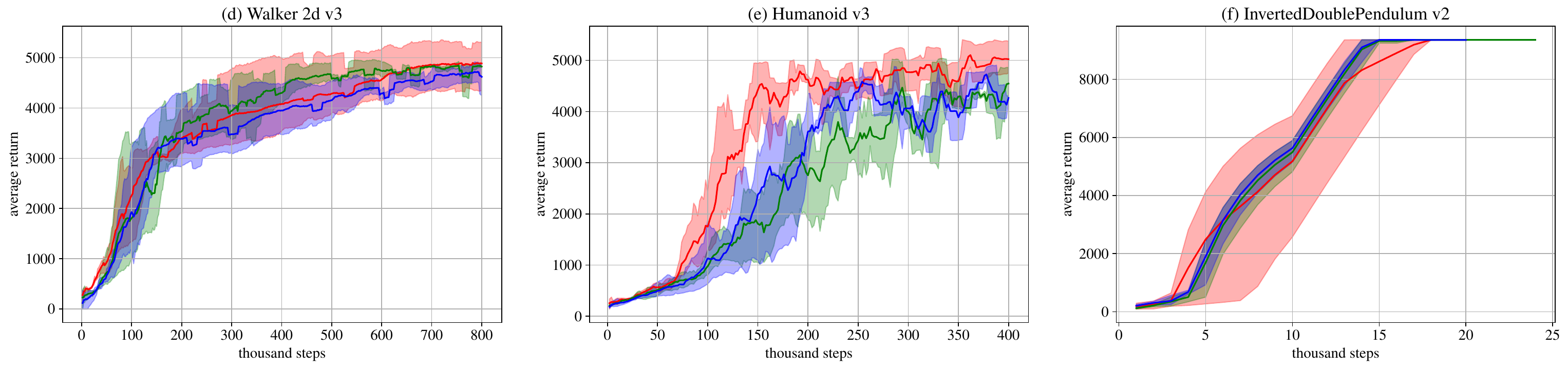}
\includegraphics[width=17.5cm]{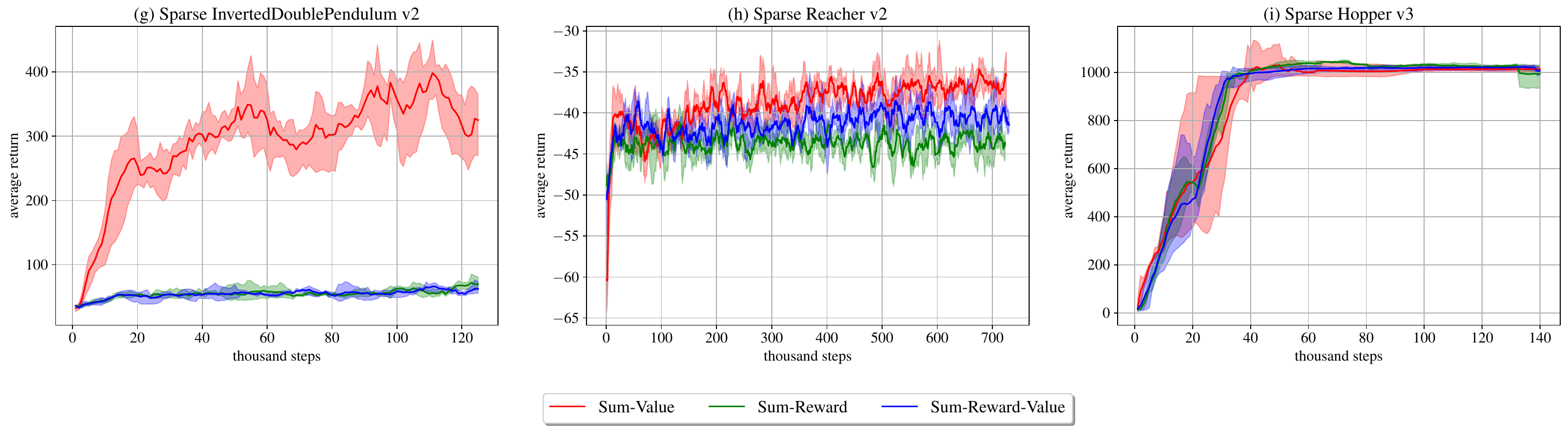}
\caption{The performance of VS-MBRL algorithm in comparison to well-known scoring functions. (i) Sum-Value: $\sum_{t'=t}^{t+H} \gamma^{t'}Q^{\pi}(s_{t},a_{t})$ (ii) Sum-Reward: $\sum_{t'=t}^{t+H} \gamma^{t'}r^{\pi}(s_{t},a_{t})$ and (iii) Sum-Reward-Value $\sum_{t'=t}^{t+H-1} \gamma^{t'}r^{\pi}(s_{t},a_{t}) + \gamma^{H}Q^{\pi}(s_{H},a_{H})$}
\label{fig4}
\end{center}
\end{figure*}
\subsection{Standard Gym Environments}
We evaluate VS-MBRL in the simulated control tasks of MuJoCo \cite{c15} included in the OpenAI-gym library \cite{c16}: Ant-v3, Hopper-v3, HalfCheetah-v3, Walker2d-v3, Humanoid-v3, and InvertedDoublePendulum-v2. In addition, to show the superior performance of our algorithm against Sum-Reward and Sum-Reward-Value scoring functions for problems with sparse rewards, we create three sparse-reward environments named Sparse InvertedDoublePendulum-v2, Sparse Reacher-v2, and Sparse Hopper-v3 by modifying the Gym environments InvertedDoublePendulum-v2, Reacher-v2, Hopper-v3, respectively. To make the rewards of an environment sparse, we only consider the reward of the last step of an episode and assume that other rewards are zero. However, it should be noted that to evaluate the sparse-reward environments, we use the corresponding dense-reward ones since the dense-reward environments are more suitable to show the performance of the algorithms.\\
We compare the average return of VS-MBRL over three trials against other scoring functions. As shown in Fig.~\ref{fig4}, the average return of each scoring function is plotted against the time step for each environment. In all environments, except Half-Cheetah-v3 and InvertedDoublePendulum-v2, VS-MBRL outperforms other scoring functions, proving the benefit of using value summation over other methods (\textit{See} Table ~\ref{tab:table1}). It is speculated that in Half-Cheetah-v3, the estimation of the state-action value function imposes a significant error in planning, especially at the first 200,000 time steps, when the state-action value neural network needs more data. Also, rewards are more useful in simple problems, such as InvertedDoublePendulum-v2, since the state-action value network requires a significant amount of training data. However, these cases are rare in complex problems, such as Humanoid-v3, Walker2d-v3, and Hopper-v3, where value summation results in fast and stable learning. According to Ant-v3 and HalfCheetah-v3, relying only on reward summation can impede learning for a long time, resulting in poor performance. Moreover, for sparse-reward environments, VS-MBRL has an excellent performance against other scoring functions, proving that relying on state-action values rather than rewards results in a better average return and sample efficiency. In addition, the performance of VS-MBRL in Sparse InvertedDoublePendulum-v2 and Sparse Reacher-v2 shows that the proposed algorithm accelerates learning in simple and complex sparse-reward environments. It is worth mentioning that, although utilizing VS-MBRL increases average returns in most environments, it may impose variance in results, which is a result of using neural networks to estimate state-action values.
\begin{table}
  \begin{center}
    \caption{Performance of Scoring Functions in Different Gym Environments}
    \label{tab:table1}
    \begin{tabular}{ |c|c|c|c| }
    \hline
      \textbf{Environments} & \textbf{Sum-Val} & \textbf{Sum-R-Val} & \textbf{Sum-R}\\
      \hline
      Ant-v3 & \textbf{7284 $\pm$ 183} & 7177 $\pm$ 56 & 289 $\pm$ 41\\
      Hopper-3 & \textbf{3427 $\pm$ 59} & 3227 $\pm$ 206 & 2507 $\pm$ 282\\
      HalfCheetah-v3 & 13839 $\pm$ 689 & \textbf{14659 $\pm$ 382} & 3744 $\pm$ 1649\\
      Walker2d-v3 & \textbf{4887 $\pm$ 409} & 4621 $\pm$ 241 & 4826 $\pm$ 6\\
      Humanoid-v3 & \textbf{5023 $\pm$ 353} & 4265 $\pm$ 41 & 4547 $\pm$ 314\\
      Inv.D.Pendul.-v2 & \textbf{9355 $\pm$ 3.5} & 9347 $\pm$ 0.1 & 9355 $\pm$ 1.8\\
      S. Inv.D.Pendul.-v2 & \textbf{325 $\pm$ 39} & 62 $\pm$ 10.9 & 69 $\pm$ 11.2\\
      S. Reacher-v2 & \textbf{-35.4 $\pm$ 2.8} & -41.4 $\pm$ 1.2 & -43.5 $\pm$ 1.3\\
      S. Hopper-v3 &  \textbf{1014 $\pm$ 12.2} & 1006 $\pm$ 16.4 & 995 $\pm$ 36.4\\
      \hline
    \end{tabular}
  \end{center}
\end{table}
\subsection{Cassie's Learning Locomotion Skills}
\begin{figure}
\begin{center}
\includegraphics[width=8.5cm]{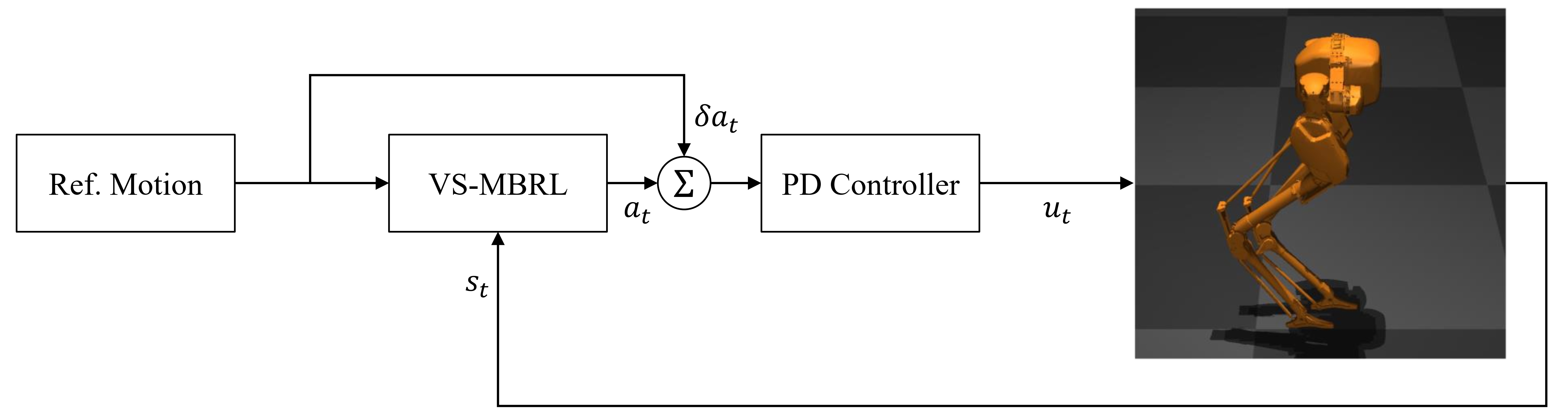}
\caption{The application of value summation model-based reinforcement learning algorithm in feedback controller. Reference motions from the expert policy are fed into the algorithm and added to the output.}
\label{fig3}
\end{center}
\end{figure}

\begin{figure}
\begin{center}
\includegraphics[width=8.5cm]{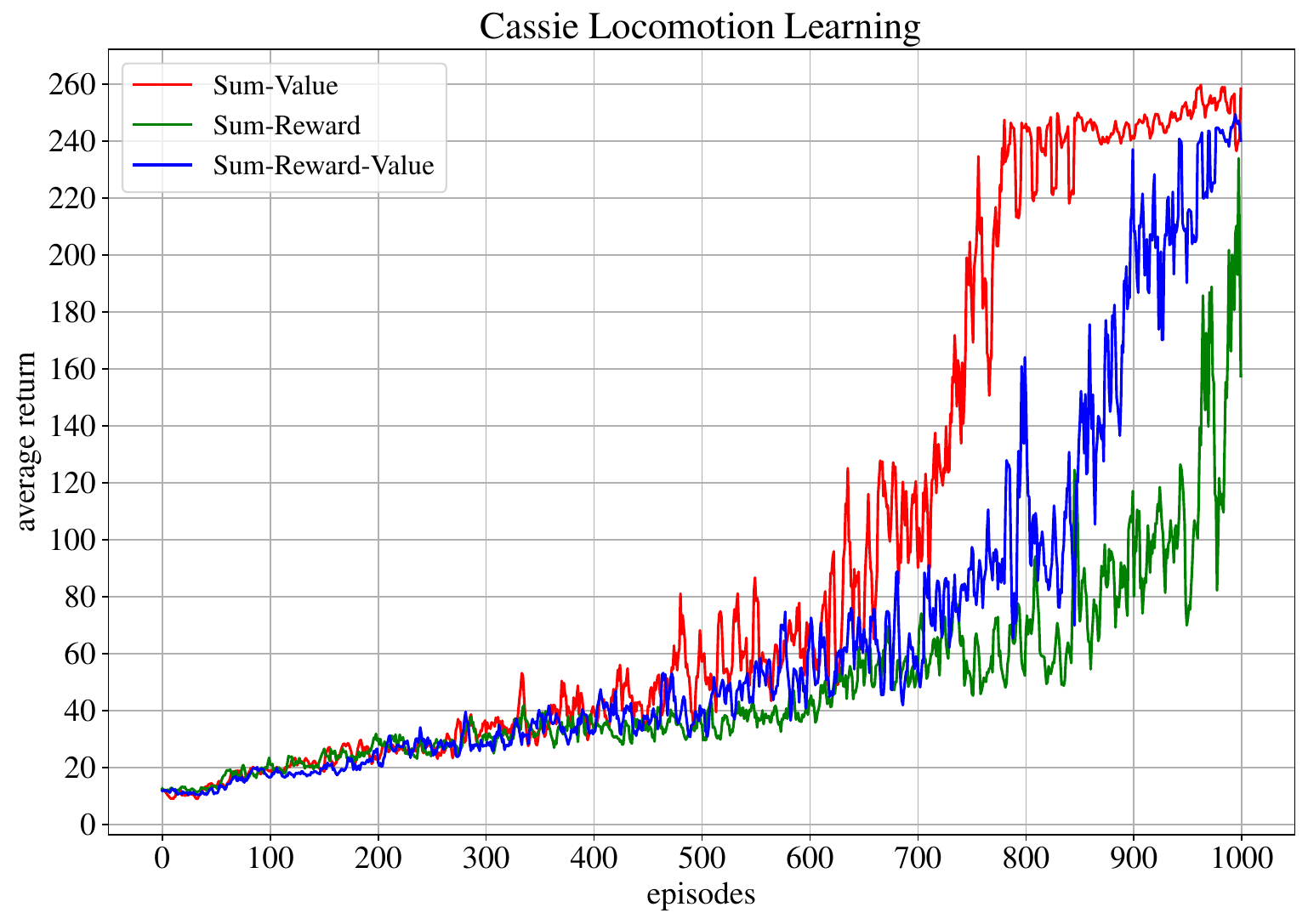}
\caption{Value Summation MBRL algorithm (Sum-Value) outperforms other scoring functions in terms of sample efficiency and average return}
\label{fig5}
\end{center}
\end{figure}

Additionally, we show that VS-MBRL is more effective than other scoring functions for the learning of Cassie's locomotion skills. Initially designed and built by Agility Robotics, the Cassie robot is approximately one meter tall and weighs 33 kg. Most of Cassie's mass is centered on the pelvis and each leg has two leaf springs to make it more flexible. Traditional controller design becomes more challenging as a result of underactuation.\\
Xie et al. \cite{c13} propose an iterative RL algorithm to increase the stability of locomotion skills through a policy distillation mechanism. In this way, the trained policy is used in the next iteration as an expert $\pi_{e}$. The expert's action is then added to that of the behavioral policy $\pi_{\theta}$ and fed into the PD controller. Each joint is controlled by a PD control loop, executed at the total rate of 2 kHz, with the targets updated every 30 ms from the policy. We combine VS-MBRL with the controller proposed in \cite{c17}, in which VS-MBRL is incorporated into the feedback control loop (Fig. ~\ref{fig3}). The controller, therefore, contains an internal loop in which the behavioral policy generates trajectories and scores them. Following this, the optimized action $a_{t}$ is summed up with the expert policy's action $\delta a$ and fed into the PD controller \cite{c17}. It was decided to set the horizon length $H$ and the number of trajectories $N$ to 3 to reduce computation time.\\
Figure ~\ref{fig5} compares the performance of VS-MBRL with traditional scoring functions. In the first 400 episodes, there is a negligible difference in performance. Throughout the next 400 episodes, the average return exponentially increases until it reaches a maximum of 250. In comparison to other Gym environments, there is a substantial performance gap between scoring functions. This is mainly because, in the simulated model of the Cassie robot for learning locomotion skills, the sampling time is $0.5$ millisecond, which is smaller than that in gym environments (Humanoid's sampling time is $3$ millisecond), thus reducing the variation of states and rewards along each trajectory, resulting in difficult reward-based planning.
\section{Conclusion}
This paper presented a Model-based RL (MBRL) algorithm, named Value Summation MBRL (VS-MBRL), which incorporates discounted value summation into planning. In addition, Soft Actor-Critic (SAC) was selected as the policy optimization algorithm, in which actions were directly sampled from a behavioral policy. We compared the proposed scoring function with two popular planning methods: discounted sum of rewards and the discounted sum of rewards added with the estimated value of the terminal state. The performance over some standard baselines indicated the superiority of the proposed method. VS-MBRL is also applied to learning locomotion skills for a simulated model of the Cassie robot. VS-MBRL proves to be a preferable alternative to current scoring functions, offering considerably greater returns for Cassie's locomotion skills as well as most MuJoCo Gym environments.
\section*{APPENDIX}
A. \textit{Proof of \eqref{eq8}}. Consider the case that we have an MDP with a finite horizon $T$ and a finite planning horizon $H$. For simplicity, we denote $Q^\pi(s_t,a_t)$ and $r(s_t,a_t)$ as $Q_t$ and $r_t$, respectively. Using \eqref{eq7} iteratively, we can expand the state-action values of time steps within the planning horizon as follows:
\begin{equation}
\begin{aligned}
\label{eq20}
Q_H & = r_H + \gamma Q_{H+1}\\
Q_{H-1} & = r_{H-1} + \gamma r_{H} + \gamma^2 Q_{H+1}\\
& \vdots\\
Q_1 & = r_1 + \gamma r_2 + \gamma^2 r_3 + ... + \gamma^{H-1} r_{H} + \gamma^H Q_{H+1}\\
Q_0 & = r_0 + \gamma r_1 + \gamma^2 r_2 + ... + \gamma^H r_H + \gamma^{H+1} Q_{H+1}
\end{aligned}
\end{equation}
The scoring function in \eqref{eq6} can be expanded as follows:
\begin{equation}
\label{eq21}
S = Q_0 + \gamma Q_1 + ... + \gamma^{H-1} Q_{H-1} + \gamma^H Q_H
\end{equation}
Substituting the state-action values from \eqref{eq20} in \eqref{eq21}, we have:
\begin{equation}
\begin{aligned}
\label{eq22}
S & = r_0 + \gamma r_1 + \gamma^2 r_2 + ... + \gamma^H r_H + \gamma^{H+1} Q_{H+1} \\
& + \gamma r_1 + \gamma^2 r_2 + \gamma^3 r_3 + ... + \gamma^{H} r_{H} + \gamma^{H+1} Q_{H+1} \\
& \vdots\\
& + \gamma^{H-1} r_{H-1} + \gamma^H r_{H} + \gamma^{H+1} Q_{H+1}\\
& + \gamma^H r_H + \gamma^{H+1} Q_{H+1}\\
& = r_0 + 2 \gamma r_1 + ... + (H+1) \gamma^H r_H + (H+1) \gamma^{H+1} Q_{H+1}\\
& = \sum_{t=0}^{H} (t+1)\gamma^{t}r_t + (H+1) \gamma^{H+1} Q_{H+1}
\end{aligned}
\end{equation}
Moreover, using the definition of the state-action value, we can write the $Q_{H+1}$ as follows:
\begin{equation}
\label{eq23}
Q_{H+1} = \sum_{t=H+1}^{T} \gamma^{t-H-1}r_t
\end{equation}
Finally, after substituting \eqref{eq23} in \eqref{eq22}, we obtain the scoring function in terms of the immediate rewards as follows:
\begin{equation}
\label{eq24}
S = \sum_{t=0}^{H} (t+1)\gamma^{t}r_t + \sum_{t=H+1}^{T} (H+1) \gamma^{t}r_t
\end{equation}

B. \textit{Proof of the Time Series}. From mathematics, we have the following time series with its solution:
\begin{equation}
\label{eq25}
\sum_{t=0}^{\infty}{\gamma^{t}} = \frac{1}{1-\gamma}
\end{equation}
The derivative of the time series in \eqref{eq25} with respect to $\gamma$ results in another time series as follows:
\begin{equation}
\label{eq26}
\begin{aligned}
\frac{d}{d\gamma} \sum_{t=0}^{\infty}{\gamma^{t}} & = \frac{d}{d\gamma} (1 + \gamma + \gamma^2 + \gamma^3 + ...) \\
& = 1 + 2 \gamma + 3 \gamma^2 + ... \\
& = \sum_{t=0}^{\infty}{(t+1)\gamma^{t}}
\end{aligned}
\end{equation}
Thus, if we take the derivative of both sides of \eqref{eq25} with respect to $\gamma$, the following relation is obtained:
\begin{equation}
\sum_{t=0}^{\infty}{(t+1)\gamma^{t}} = \frac{1}{(1-\gamma)^{2}}
\end{equation}



\printbibliography[heading=bibintoc]
\end{document}